\def\year{2022}\relax
\documentclass[letterpaper]{article} 
\usepackage{aaai22}  
\usepackage{times}  
\usepackage{helvet,xcolor}  
\usepackage{courier}  
\usepackage[hyphens]{url}  
\usepackage{graphicx} 
\usepackage{subfigure}
\usepackage{gensymb}
\usepackage{natbib}  
\usepackage{caption} 
\DeclareCaptionStyle{ruled}{labelfont=normalfont,labelsep=colon,strut=off} 
\usepackage{algorithm}
\usepackage{algorithmic}
\usepackage{enumitem,booktabs}
\usepackage{newfloat}
\usepackage{listings}
\floatstyle{ruled}
\newfloat{listing}{tb}{lst}{}
\floatname{listing}{Listing}
\usepackage{color}
\definecolor{yamabuki}{RGB}{248, 181, 0}

\title{Demystifying the Data Need of ML-surrogates for CFD Simulations}
\author {Tongtao Zhang\footnote{Equal Contribution}, Biswadip Dey$^*$, Krishna Veeraraghavan$^*$, Harshad Kulkarni, Amit Chakraborty}
\affiliations{Siemens Corporation\\
\{tongtao.zhang, biswadip.dey, krishna.veeraraghavan, harshad.kulkarni1, 	amit.chakraborty\}@siemens.com}

\begin{document}
\maketitle
%
%
%
\begin{abstract}
Computational fluid dynamics (CFD) simulations, a critical tool in various engineering applications, often require significant time and compute power to predict flow properties. The high computational cost associated with CFD simulations significantly restricts the scope of design space exploration and limits their use in planning and operational control. To address this issue, machine learning (ML) based surrogate models have been proposed as a computationally efficient tool to accelerate CFD simulations. However, a lack of clarity about CFD data requirements often challenges the widespread adoption of ML-based surrogates among design engineers and CFD practitioners. In this work, we propose an ML-based surrogate model to predict the temperature distribution inside the cabin of a passenger vehicle under various operating conditions and use it to demonstrate the trade-off between prediction performance and training dataset size. Our results show that the prediction accuracy is high and stable even when the training size is gradually reduced from 2000 to 200. The ML-based surrogates also reduce the compute time from $\sim$30 minutes to around $\sim$9 milliseconds. Moreover, even when only 50 CFD simulations are used for training, the temperature trend (e.g., locations of hot/cold regions) predicted by the ML-surrogate matches quite well with the results from CFD simulations.
\end{abstract}
%
%
%

\section{Introduction}

The existing approaches for designing complex devices and systems in the automotive industry, for example, aerodynamic surfaces and thermal, energy, and battery management systems, typically involve an iterative interaction between exploration of the design/operating space and evaluation of the performance using high-fidelity CFD simulations. However, the computational cost associated with high-fidelity CFD simulations using commercial solvers, for example, SimCenter Star-CCM+ or Ansys Fluent, severely limits the scope of the overall design optimization process and may, as a consequence, lead to sub-optimal design choices. From this perspective, using neural networks to capture the relevant functional relationship to predict simulation outcomes and applying ML algorithms to develop a fast and accurate surrogate for CFD simulations have the potential to significantly accelerate design evaluations and thereby generate improved design choices.
\begin{figure}[t!]
\centering
\includegraphics[width=0.49\textwidth]{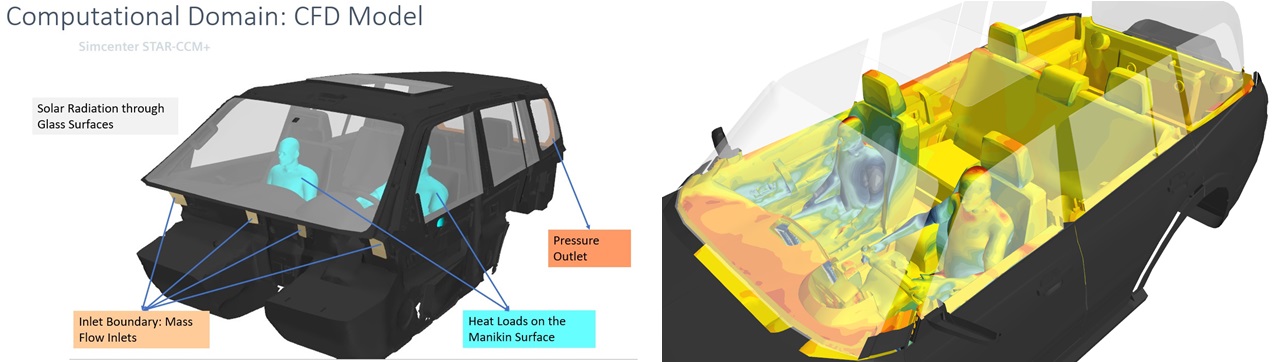}
\small{\caption{The design case under consideration. The left figure shows the configuration of the car cabin along with the relevant input variables and the right figure shows the temperature profile on the solution plane of our interest.}\label{fig:CFDModel}}

\vspace{-2em}
\end{figure}

Over the last decade, artificial intelligence and data-driven approaches have transformed the state-of-the-art in almost every industry. In a variety of application domains, such as image recognition \cite{He_2016_CVPR}, sequential decision making \cite{mnih2015humanlevel}, and language comprehension \cite{BERT_paper}, deep neural networks and representation learning \cite{goodfellow2016deep} have become widely-adopted as the enabler of state-of-the-art solution approaches. As ML-based approaches can help reduce costs and create new differentiated values, applying them to CFD applications can be a strategic asset to companies. ML-based solutions offer substantial benefits to CFD engineers, designers, and analysts. In addition to reducing computational, design program, and operational costs by creating more design insight per simulation at a faster turnaround time, these approaches can also reduce the process and program development time with ML based surrogate models and smart AI driven workflows and thereby expedite the turnaround time.

Use of ML-based approaches to accelerate CFD simulations has drawn significant attention over the last few years. A fast growing body of work \cite{hennigh2017lat, jiang2020meshfreeflownet, nabian2020physics, raissi2019physics, wang2020towards, warey2020data, white2019fast} has demonstrated that ML-based approaches can predict simulation results, i.e., flow and temperature profiles over a spatio-temporal domain, in both mesh-based and mesh-free manner and incorporation of physics-based regularization in these formulations improves the quality of results by a significant margin. Prior work has also introduced appropriate Fourier features to incorporate relevant periodic behavior in the predicted solutions \cite{zhang2020frequencycompensated, li2021fourier}. On the other hand, alternative approaches \cite{dwivedi2019distributed, lu2019deepxde, nabian2019deep} use self-supervised learning to predict solutions of the underlying partial differential equation and thereby avoid the need for the computationally expensive step of dataset generation.

Although the existing approaches primarily focus on prediction accuracy, they provide little to no insight into the trade-off between data availability and prediction accuracy. However, from the perspective of a CFD practitioner, it is very critical to know how the prediction accuracy of an ML-based surrogate varies with changes in the dataset size. We provide an answer to this question in this work. The main contribution of this work is two-fold: -- first, we introduce a neural network based surrogate model that uses very few CFD simulations to predict temperature profile inside a car cabin; and second, we study the impact of training dataset size on prediction accuracy and show that reasonably accurate surrogate models can be learned even with a tiny dataset.
\section{Problem and Framework}
\subsection{Problem}
In this paper, we focus on the problem of predicting temperature distribution inside the cabin for various operating conditions. As shown in Figure~\ref{fig:CFDModel}, we are working on a closed volume of the passenger compartment with two manikins in the front seat. We excluded the conductive heat transfer between the manikin and the seat to simplify the CFD model. As a result, the model only considers the convective and the radiative modes of heat transfer and takes into account the effects from convective heat transfer (between the air introduced in the passenger compartment from the HVAC system and the manikins) and radiative heat transfer (including the solar radiation between the cabin's glazing surfaces and other surrounding walls and the manikins). In this formulation, we focus on the following six variables which influence the temperature profile: Solar load $\in$ $\{500, 600, 700, 800, 900, 1000\}W/m^2$; Sun altitude $\in$ $\{45, 90\}\degree$; Sun azimuth $\in$ $\{-90, 90\}\degree$; Discharge air temperature $\in$ $\{5, 10, 15\}\degree C$; Volume flow rate $\in$ $\{50, 100, 150, 200, 250, 300\}CFM$; and Ambient temperature $\in$ $\{20, 25, 30, 35, 40\}\degree C$. 
%

Then we employ the standard $k-\epsilon$ turbulence model and use a commercial CFD tool (Simcenter STAR-CCM+ v15.04.010) to simulate and visualize the steady state temperature distribution under each of the input combinations. However, even with the simplifying assumptions and aforementioned constraints, designer and engineers still need significant compute power and time to simulate and visualize each and every combination of the input variable. This hinders fast prototyping and agile design cycles given limited time and resource. 

\subsection{Learning Framework}
\begin{figure}[t!]
\centering
\includegraphics[width=0.45\textwidth]{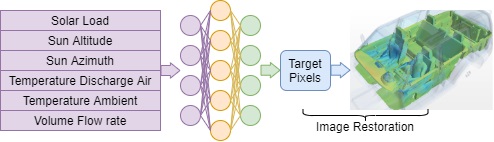}
\small{\caption{Pipeline of the proposed ML framework.}\label{fig:pipeline}}
\vspace{-1.7em}
\end{figure}

As shown in Figure \ref{fig:pipeline}, we take the six variables as the input and feed them into a multilayer perceptron (MLP). Each layer in the neural network is a fully-connected layer with a hyperbolic tangent activation function:
\begin{equation}
    y_i = \tanh(FC_i(x_i)),
\end{equation}
where $x$ is the input of each layer, $y$ is the output and $i$ is the layer index. The MLP outputs the ``target pixels'' whose values indicate the temperature in the visualization. We use mean square error (MSE) as the loss function to measure the difference between the prediction and the ground truth; and we take Adam Optimizer~\cite{kingma2014adam} to backpropagate and train the neural network.

\begin{figure*}[t!]
\centering
\subfigure[A low temperature example wherein the strong ($300CFM$) and cold ($10\degree C$) airflow overwhelms the warm ambient condition ($40\degree C$).
\label{fig:7_1320}]{\includegraphics[width=0.32\textwidth]{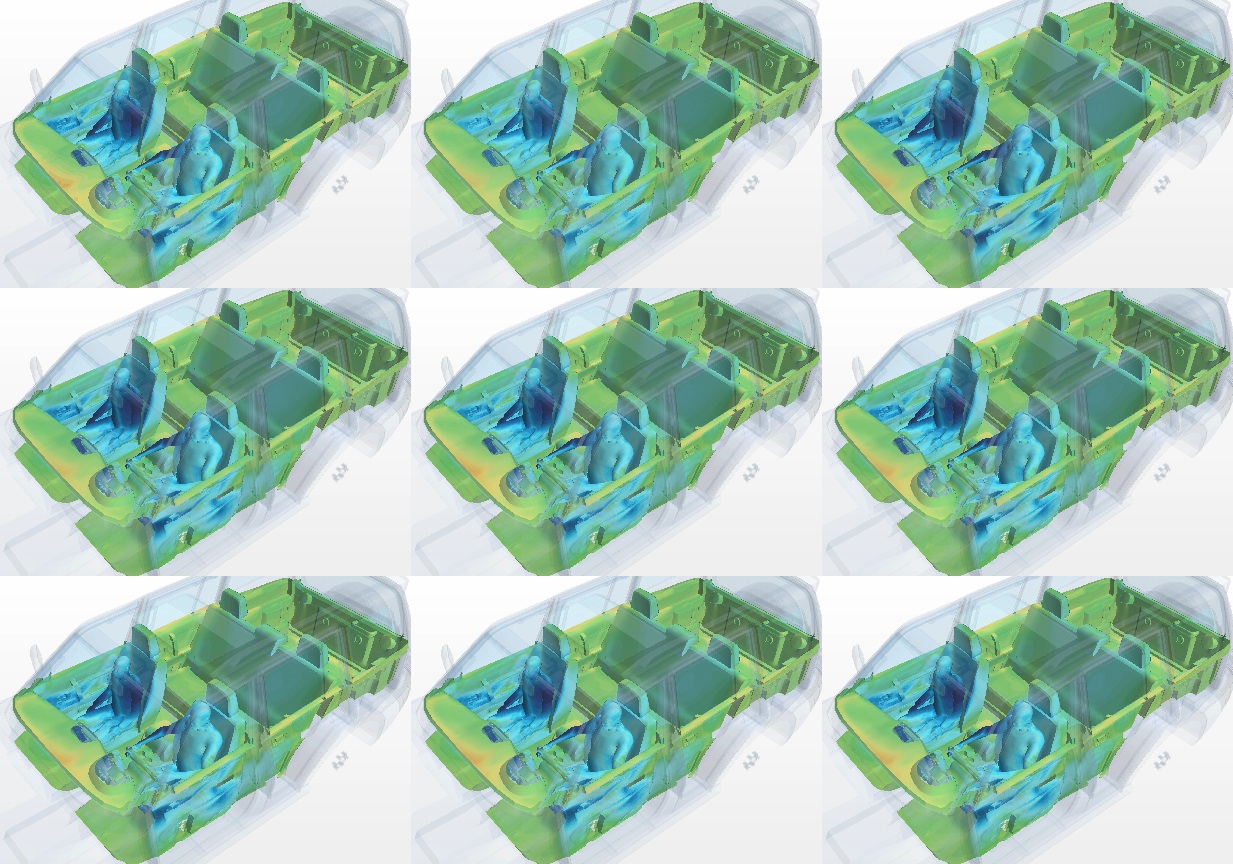}}
~
\subfigure[A high temperature example wherein the sunshine hits the driver's shoulder, and ambient temperature is a high ($35\degree C$) with very little airflow ($50CFM$).
\label{fig:7_499}]{\includegraphics[width=0.32\textwidth]{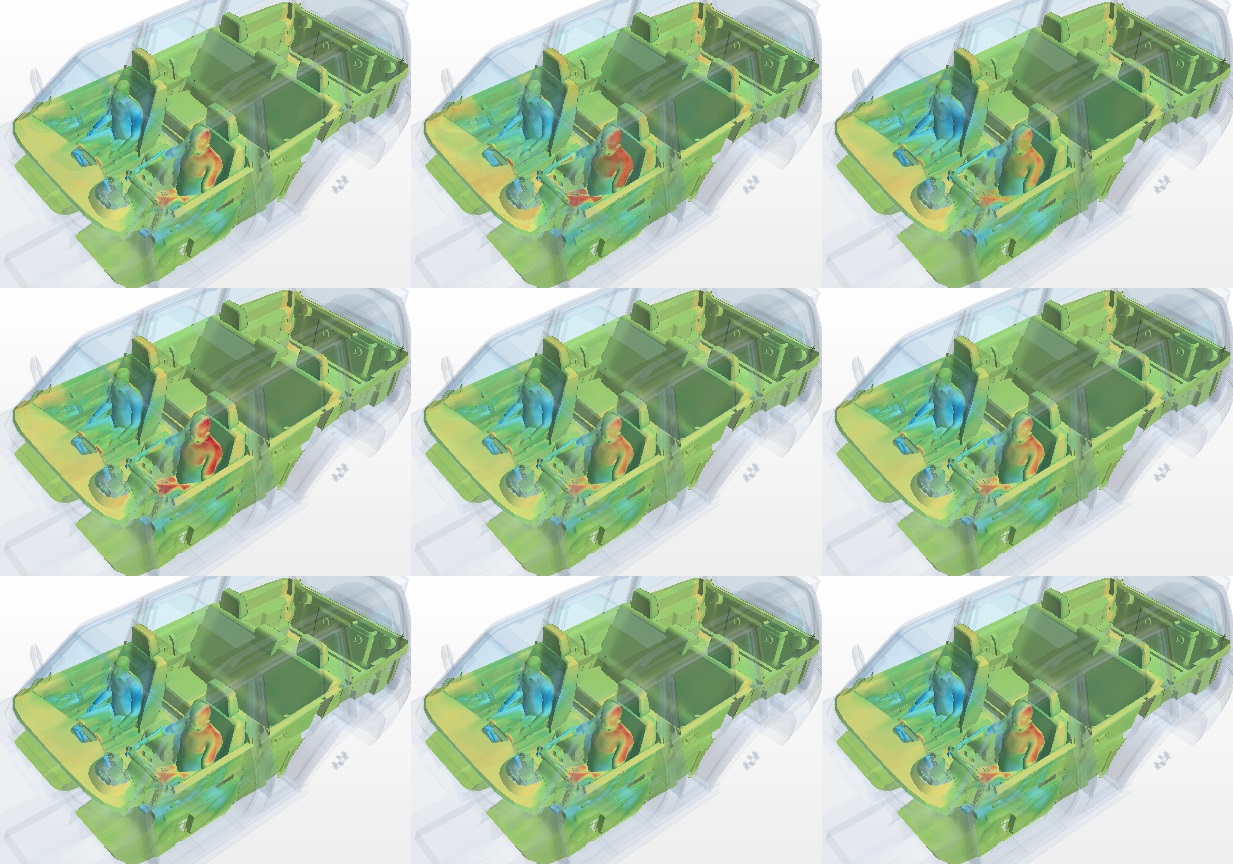}}
~
\subfigure[An intermediate temperature example with a moderate ambient temperature ($30\degree C$) and very little airflow ($50CFM$).
\label{fig:7_583}]{\includegraphics[width=0.32\textwidth]{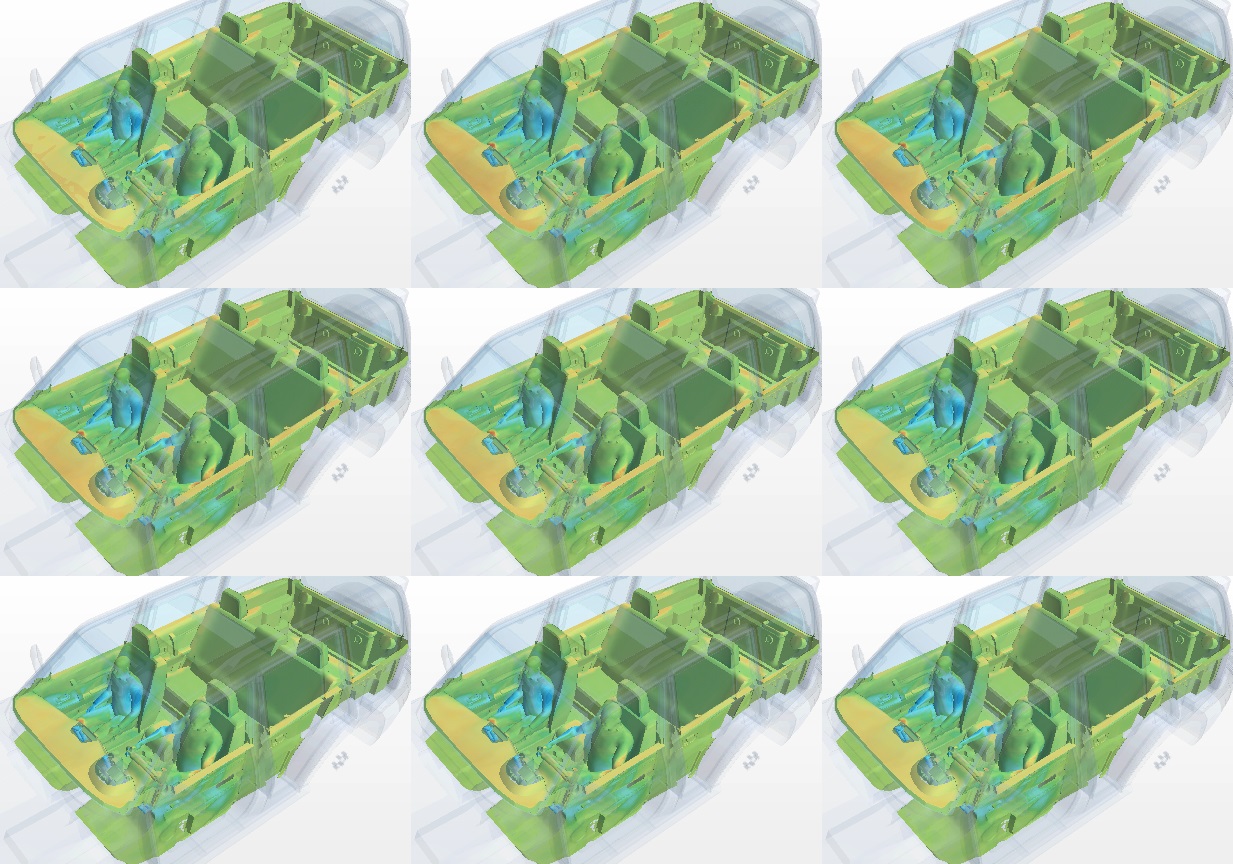}}
\vspace{-1em}
\small{\caption{Examples from test set. In each sub-figure, the ground truth on the upper left. The predictions with $50$ and $100$ training instances are on the first row, $150$, $200$ and $500$ are on the second row and $1000$, $1500$ and $2000$ on the bottom. Please refer to supplementary material for enlarged version and comparison.}\label{fig:examples}}
\vspace{-1.5em}
\end{figure*}
\section{Experiments}

\subsection{Dataset Overview and Preprocessing}
To evaluate the model, we form $2160$ distinct cases from the $6$ variables and generate simulation and visualization images for the cases. We randomly sample $2000$ cases as the training set, $80$ as validation and $80$ as test. We also inspect the three sets and ensure that all sets cover the same ranges of variables in the input.

To find the target pixels, we scan all the images and collect the coordinates of the pixels whose values change among the images. With the coordinates, the output of our model will be restored to the images to illustrate the temperature distribution.

\subsection{Hyper-parameters and models}
For all the experiments, we set learning rate as $0.001$, mini-batch size as $32$ and max epoch as $2000$. With random seeds, we initialize the network parameters. To train a model with $k$ training instances, where $k\in\{50, 100, 150, 200, 500, 1000, 1500, 2000\}$, we use the same seeds to shuffle the whole training set and sample the first $k$ instances as training data. We select the \textit{best} epoch as the epoch when the MSE on the validation set achieves its minimum value. We also use Structural Similarity (SSIM) indices \cite{Wang2004Image} to evaluate the performance on the validation and test set.

With random seed $0$, we grid-search the hyper-parameters for neural network structure. For number of layers we try $5$ and $10$, and for the size of hidden layers we try $64$, $128$, $256$. We observe SSIM indices on the test set from the best models of all 6 combinations with 8 different training sizes. We find that $10$ layers with the size of $128$ performs the best as it has 4 best SSIM indices among the 8. Hence, the results reported in the section are all based on this combination of hyper-parameters. With random seeds $1$ to $10$, we run the experiments and report the performance.

We also implement a linear model with the identical input, output, data splits and random seeds as baseline. We also report the SSIM indices for the linear model.

\subsection{Results and Discussion}
\textbf{Results and Examples:} Table~\ref{tab:error_rate} demonstrates the SSIM indices from the experiments. We conclude that our proposed method significantly outperforms the linear baseline. We observe that, even with 50 training instances, the loss is very small and the SSIM index shows that the prediction is very close to the ground truth. Increasing the size of training set brings forth further improvement in the performance. According to Figure~\ref{fig:examples}, starting from training sizes of $200$, the difference becomes barely visible.

\begin{figure}[th]
\centering
\subfigure[Ground truth]{\includegraphics[width=0.24\textwidth]{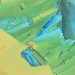}}~
\subfigure[Prediction from 50]{\includegraphics[width=0.24\textwidth]{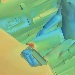}}
\vspace{-1em}
\small{\caption{Zoomed-in view of the passenger's lap from Figure~\ref{fig:7_583}}\label{fig:lap}}
\vspace{-1em}
\end{figure}

Figure~\ref{fig:lap} illustrates the amplified patches from Figure~\ref{fig:7_583}, we compare the ground truth and the result trained from 50 images. Our model is still able to capture points of interest such as the hot point on the ``right lap'' of the ``passenger'' and demonstrate the temperature trends.

\begin{figure}[th]
\centering
\subfigure[Ground truth]{\includegraphics[width=0.24\textwidth]{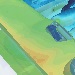}}~
\subfigure[Prediction from 200]{\includegraphics[width=0.24\textwidth]{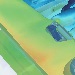}}
\vspace{-1em}
\small{\caption{Zoomed-in view of the dashboard from Figure~\ref{fig:7_1320}}\label{fig:contour}}.
\vspace{-2em}
\end{figure}

\textbf{Prediction Accuracy:} One observation is that fewer training instances lead to larger difference, higher losses and lower similarity. Another observation which explains the slight difference between ground truth and prediction is shown in Figure~\ref{fig:contour}. The ground truth has clearer ``contour'' style texture, while prediction output looks more blurred.

\textbf{Time Consumption:} With a server equipped with two Intel Xeon Gold 6226R processors (2.9GHz and 64 threads), we need $0.75$ hours for each simulation case with the CFD tool, and two months for a single server (long time) or two days for thirty machines (high cost) to simulate all $2160$ cases. We run our framework on \textbf{the same single server} without any additional hardware such as GPU/TPU because we want to ensure fair comparison and fully-connected layers are highly optimized on CPU. Table~\ref{tab:time_consuming} demonstrates the total time of simulating a number of training cases and predicting all the other cases. Prediction is almost instantaneous, so we omit it in the table. With $200$ training instances we only need 10\% of the total time and we are able to achieve comparable performance as discussed. We conclude that our proposed method offers a good solution when the availability of time and computing resource is limited.
\section{Conclusion}

\begin{table*}[t!]
\centering
\begin{tabular}{c|llllllll}
\toprule
\# of Training Instances & 50 & 100 & 150 & 200 & 500 & 1000 & 1500 & 2000
\\
\midrule
Validation SSIM & $0.724$ & $0.726$ & $0.738$ & $0.804$ & $0.930$ & $0.937$ & $0.940$ & $0.937$
\\
(baseline) & $\pm0.004$ & $\pm0.003$ & $\pm0.011$ & $\pm0.036$ & $\pm0.024$ & $\pm0.022$ & $\pm0.020$ & $\pm0.019$
\\
\midrule
Test SSIM & $0.724$ & $0.726$ & $0.739$ & $0.809$ & $0.943$ & $0.949$ & $0.952$ & $0.949$
\\
(baseline) & $\pm0.002$ & $\pm0.003 $ & $\pm0.015$ & $\pm0.035$ & $\pm0.019$ & $\pm0.018$ & $\pm0.015$ & $\pm0.014$\\
\midrule
Validation SSIM & $0.977$ & $0.984$ & $0.987$ & $0.990$ & $0.995$ & $0.996$ & $0.996$ & $0.997$
\\
(our approach) & $\pm0.012$ & $\pm0.009$ & $\pm0.008$ & $\pm0.006$ & $\pm0.002$ & $\pm0.001$ & $\pm0.001$ & $\pm0.001$
\\
\midrule
Test SSIM & $0.977$ & $0.983$ & $0.986$ & $0.989$ & $0.995$ & $0.996$ & $0.997$ & $0.997$
\\
(our approach) & $\pm0.012$ & $\pm0.010 $ & $\pm0.009$ & $\pm0.007$ & $\pm0.002$ & $\pm0.001$ & $\pm0.001$ & $\pm0.001$\\
\bottomrule
\end{tabular}
\caption{Similarity performance on the validation and test sets from the best models of the baseline and our proposed approach with regard to the training set size.}
\label{tab:error_rate}
\vspace{-1.8em}
\end{table*}


Machine learning with CFD can be a good complimentary tool in engineering design. Simulation data can be a strategic asset for the companies if they figure out how the data they generate can be leveraged with surrogate model for future data driven design decisions. Clear understanding of what type of CFD data and how much CFD data can be generated is essential for defining the goals of the surrogate model and for the successful transformation of AI journey in simulation and design persona. From our studies, we infer that with very few data samples (\textit{e.g.}, 50), we are able to predict the trend of where hot and cold temperature zones can be in the cabin; and with 500 data samples, we are able to accurately predict the magnitude of the temperature distribution in the cabin as compared to the CFD simulation. Moreover, with 50 data samples, engineers can weed out the bad designs using machine learning models and then run CFD simulation for more streamlined design exploration studies; and with 500 data samples engineers can skip the CFD simulations at the early design or conceptual design phase and use CFD for more challenging physics and modelling problems at detailed design stage.
\begin{table}[h!]
  \centering
  \begin{tabular}{c|ccc|c}
    \toprule
    Dataset & Simulation & Training & Total & Reduction
    \\
    Size & Time & Time & Time & (\%)
    \\
    \midrule
    50  & 37.5 & 0.40 & 37.90 & 97.66\\
    100 & 75 & 0.54 & 75.54 & 95.34\\
    150 & 112.5 & 0.60 & 113.10 & 93.02\\
    200 & 150 & 0.66 & 150.66 & 90.70\\
    500 & 375 & 0.99 & 375.99 & 76.79\\
    1000 & 750 & 1.58 & 751.58 & 53.61\\
    1500 & 1125 & 2.18 & 1127.18 & 30.42\\
    2000 & 1500 & 2.91 & 1502.91 & 7.23\\
    \bottomrule
    \end{tabular}
  \caption{Time consumption in hours and percentage reduction in compute time.}
  \label{tab:time_consuming}
\end{table}


{\small{
\bibliography{refs}
}}

\end{document}